\documentclass[11pt]{article}
\usepackage{graphicx}    % needed for including graphics e.g. EPS, PS
\usepackage{amsmath}
\usepackage{amsfonts}
\usepackage{amssymb}
\usepackage{bbm}

\begin{document}         
% Start your text
% don't want date printed
\date{}

% >>>>>>>>>>>>>>>>>>>>>>>  Put your title here <<<<<<<<<<<<<<<<<<<<<<<<
% make title bold and 14 pt font (Latex default is non-bold, 16pt) 
\title{\Large {\bf A Gentle Tutorial of Recurrent Neural Network with Error Backpropagation}}
\author{Gang Chen 
\thanks{email comments to gangchen@buffalo.edu}
\\
Department of Computer Science and Engineering, SUNY at Buffalo
}
\maketitle

\section{abstract}
We describe recurrent neural networks (RNNs), which have attracted great attention on sequential tasks, such as handwriting recognition, speech recognition and image to text. However, compared to general feedforward neural networks, RNNs have feedback loops, which makes it a little hard to understand the backpropagation step. Thus, we focus on basics, especially the error backpropagation to compute gradients with respect to model parameters. Further, we go into detail on how error backpropagation algorithm is applied on long short-term memory (LSTM) by unfolding the memory unit.
% -----------------------------------
\section{Sequential data}
Sequential data is common in a wide variety of domains including natural language processing, speech recognition and computational biology. In general, it is divided into time series and ordered data structures. As for the time-series data, it changes over time and keeps consistent in the adjacent clips, such as the time frames for speech or video analysis, daily prices of stocks or the rainfall measurements on successive days. There are also ordered data in the sequence, such as text and sentence for handwriting recognition, and genes. For example, successfully predicting protein-protein interactions requires knowledge of the secondary structures of the proteins and semantic analysis might involve annotating tokens with parts of speech tags.

The goal of this work is for sequence labeling, i.e. classify all items in a sequence \cite{Chen16}. For example, in handwritten word recognition we wish to label a sequence of characters given features of the characters; in part of speech tagging, we wish to label a sequence of words.

More specifically, given an observation sequence ${\bf x} = \{ {\bf x}_1, {\bf x}_2,..., {\bf x}_T  \}$ and its corresponding label $y = \{ y_1, y_2,..., y_T   \}$, we want to learn a map $f: {\bf x}  \mapsto y$. In the following, we will introduce RNNs to model the sequential data and give details on backpropagation over the feedback loop. 

\begin{figure*}
\centering
\begin{tabular}{c}
\includegraphics[trim=10 130 10 100, clip, width=12cm]{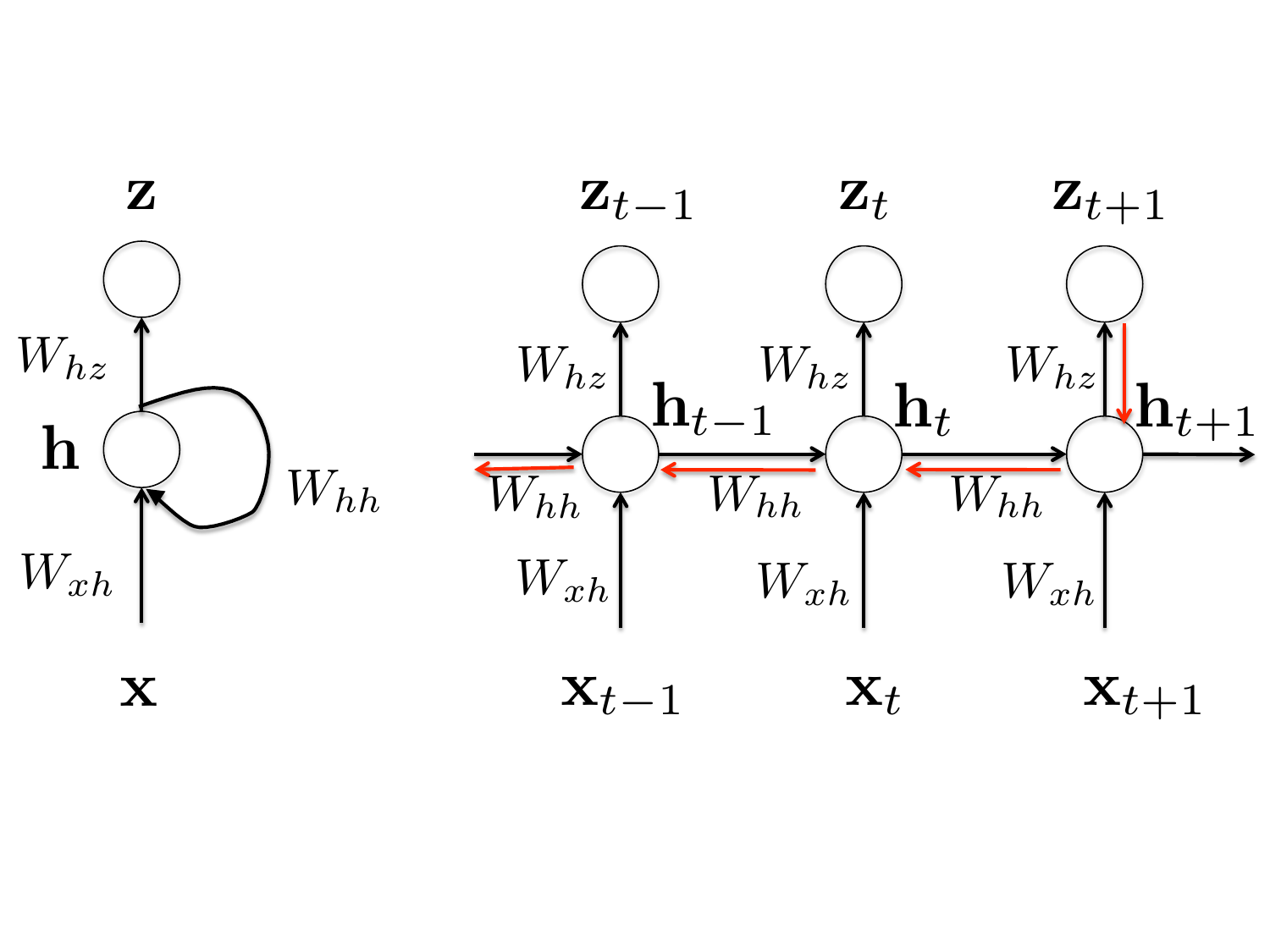} 
\end{tabular}
\caption{It is a RNN example: the left recursive description for RNNs, and the right is the corresponding extended RNN model in a time sequential manner.}
\label{ch6-fig:rnn_example}
\end{figure*}
\section{Recurrent neural networks}
A RNN is a kind of neural networks, which can send feedback signals (to form a directed cycle), such as Hopfield net \cite{Hopfield88} and long-short term memory (LSTM) \cite{Hochreiter97}. RNN models a dynamic system, where the hidden state ${\bf h}_t$ is not only dependent on the current observation ${\bf x}_t$, but also relies on the previous hidden state ${\bf h}_{t-1}$. More specifically, we can represent ${\bf h}_t$ as
\begin{align}\label{ch6-eq:rnns}
{\bf h}_t = f({\bf h}_{t-1},  {\bf x}_t)
\end{align}
where $f$ is a nonlinear mapping. Thus, ${\bf h}_t$ contains information about the whole sequence, which can be inferred from the recursive definition in Eq. \ref{ch6-eq:rnns}. In other words, RNN can use the hidden variables as a memory to capture long term information from a sequence.

Suppose that we have the following RNN model, such that
\begin{align}
& {\bf h}_t =  tanh( W_{ hh} {\bf h}_{t-1} +   W_{ xh}{\bf x}_t + {\bf b}_{\bf h}) \label{ch6-eq:rnnscase} \\
& z_t = softmax(W_{hz} {\bf h}_t + {\bf b}_z) \label{ch6-eq:rnnscase2}
\end{align}
where $z_t$ is the prediction at the time step $t$, and $tanh(x)$ is defined as
\begin{align}
tanh(x) = \frac{ sinh(x) }{ cosh(x)  } = \frac{ e^x - e^{-x} }{  e^x+ e^{-x}    } = \frac{  e^{2x} -1 }{ e^{2x} +1 }  \nonumber
\end{align}

More specifically, the RNN model above has one hidden layer as depicted in Fig. \ref{ch6-fig:rnn_example}. Notice that it is very easy to extend the one hidden case into multiple layers, which has been discussed in deep neural network before. Considering the varying length for each sequential data, we also assume the parameters in each time step are the same across the whole sequential analysis. Otherwise it will be hard to compute the gradients. In addition, sharing the weights for any sequential length can generalize the model well. As for sequential labeling, we can use the maximum likelihood to estimate model parameters. In other words, we can 
minimize the negative log likelihood the objective function 
\begin{align}\label{ch6-eq:rnnmaxfun}
\mathcal{L}({\bf x}, {\bf y}) = -\sum_t y_t log z_t
\end{align}
In the following, we will use notation $\mathcal{L}$ as the objective function for simplicity. And further we will use $\mathcal{L}(t+1)$ to indicate the output at the time step $t+1$, s.t. $\mathcal{L}(t+1) = - y_{t+1} log z_{t+1} $. 

Let's set $\alpha_t = W_{hz} {\bf h}_t + {\bf b}_z$, and then we have $z_t = softmax(\alpha_t )$ according to Eq. \ref{ch6-eq:rnnscase2}. By taking the derivative with respect to $\alpha_t$ (refer to appendix for details), we get the following
\begin{align}\label{ch6-eq:rnngradz}
\frac{ \partial \mathcal{L} }{ \partial \alpha_t } = -(y_t- z_t)
\end{align}
Note the weight $W_{hz}$ is shared across all time sequence, thus we can differentiate to it at each time step and sum all together
\begin{align}\label{ch6-eq:rnngradwhz}
\frac{ \partial \mathcal{L} }{ \partial W_{hz} } = \sum_t \frac{ \partial \mathcal{L} }{ \partial z_t } \frac{\partial z_t}{ \partial W_{hz} }
\end{align}

Similarly, we can get the gradient w.r.t. bias $b_{z}$
\begin{align}\label{ch6-eq:rnngradbz}
\frac{ \partial \mathcal{L} }{ \partial b_{z} } = \sum_t \frac{ \partial \mathcal{L} }{ \partial z_t } \frac{\partial z_t}{ \partial b_{z} }
\end{align}

Now let's go through the details to derive the gradient w.r.t. $W_{hh}$. Considering at the time step $t \rightarrow t+1$ in Fig. \ref{ch6-fig:rnn_example},  
\begin{align}\label{ch6-eq:rnngradwhh1}
\frac{ \partial \mathcal{L}(t+1)  }{ \partial W_{hh} } = \frac{ \partial \mathcal{L}(t+1) }{  \partial z_{t+1}  }    \frac{  \partial z_{t+1} }{\partial {\bf h}_{t+1}  }   \frac{ \partial {\bf h}_{t+1}  }{ \partial W_{hh} } 
\end{align}
where we only consider one step $t \rightarrow (t+1)$. And because the hidden state ${\bf h}_{t+1}$ partially dependents on ${\bf h}_{t}$, so we can use backpropagation to compute the above partial derivative. Think further $W_{hh}$ is shared cross the whole time sequence, according to the recursive definition in Eq. \ref{ch6-eq:rnnscase}. Thus, at the time step $(t-1) \rightarrow t$, we can further get the partial derivative w.r.t. $W_{hh}$ as follows
\begin{align}\label{ch6-eq:rnngradwhh2}
\frac{ \partial \mathcal{L}(t+1) }{ \partial W_{hh} } =  \frac{ \partial \mathcal{L}(t+1) }{  \partial z_{t+1}  }    \frac{  \partial z_{t+1} }{ \partial {\bf h}_{t+1}  }   \frac{ \partial {\bf h}_{t+1} }{ \partial {\bf h}_t  }  \frac{  \partial {\bf h}_{t}  }{ \partial W_{hh} } 
\end{align}

Thus, at the time step $t+1$, we can compute gradient w.r.t. $z_{t+1}$ and further use backpropagation through time (BPTT) from $t$ to $0$ to calculate gradient w.r.t. $W_{hh}$, shown as the red chain in Fig. \ref{ch6-fig:rnn_example}. Thus, if we only consider the output $z_{t+1}$ at the time step $t+1$, we can yield the following gradient w.r.t. $W_{hh}$
\begin{align}
\frac{ \partial \mathcal{L}(t+1) }{ \partial W_{hh} } =\sum_{k=1}^t  \frac{ \partial \mathcal{L}(t+1) }{  \partial z_{t+1}  }    \frac{  \partial z_{t+1} }{\partial {\bf h}_{t+1}  }   \frac{ \partial{\bf h}_{t+1} }{ \partial {\bf h}_k  }  \frac{ \partial {\bf h}_{t}  }{ \partial W_{hh} } 
\end{align}
Aggregate the gradients w.r.t. $W_{hh}$ over the whole time sequence with back propagation, we can finally yield the following gradient w.r.t. $W_{hh}$
\begin{align}\label{ch6-eq:rnngradwhh}
\frac{ \partial \mathcal{L} }{ \partial W_{hh} } = \sum_t \sum_{k=1}^{t+1}  \frac{ \partial \mathcal{L}(t+1) }{  \partial z_{t+1}  }    \frac{  \partial z_{t+1} }{ \partial {\bf h}_{t+1}  }   \frac{  \partial {\bf h}_{t+1} }{ \partial {\bf h}_k  }  \frac{   \partial {\bf h}_{k}  }{ \partial W_{hh} } 
\end{align}

Now we turn to derive the gradient w.r.t. $W_{xh}$. Similarly, we consider the time step $t+1$ (only contribution from ${\bf x}_{t+1}$) and calculate the gradient w.r.t. to $W_{xh}$ as follows
\begin{align}
\frac{ \partial \mathcal{L}(t+1) }{ \partial W_{xh} } = \frac{ \partial \mathcal{L}(t+1) }{ \partial {\bf h}_{t+1} }  \frac{ \partial {\bf h}_{t+1}  }{  \partial W_{xh}  }  \nonumber
\end{align}
Because ${\bf h}_t$ and ${\bf x}_{t+1}$ both make contribution to ${\bf h}_{t+1}$, we need to backpropagte to ${\bf h}_t$ as well. If we consider the contribution from the time step $t$, we can further get 
\begin{align}
& \frac{ \partial \mathcal{L}(t+1) }{ \partial W_{xh} } = \frac{ \partial \mathcal{L}(t+1) }{ \partial {\bf h}_{t+1} }  \frac{ \partial {\bf h}_{t+1}  }{  \partial W_{xh}  }  + \frac{ \partial \mathcal{L}(t+1) }{ \partial {\bf h}_{t} }  \frac{ \partial {\bf h}_{t}  }{  \partial W_{xh}  }  \nonumber  \\
 = & \frac{ \partial \mathcal{L}(t+1) }{ \partial {\bf h}_{t+1} }  \frac{ \partial {\bf h}_{t+1}  }{  \partial W_{xh}  }  + \frac{ \partial \mathcal{L}(t+1) }{ \partial {\bf h}_{t+1} }     \frac{ \partial {\bf h}_{t+1}  }{ \partial {\bf h}_{t}  }      \frac{ \partial {\bf h}_{t}  }{  \partial W_{xh}  } 
\end{align}

Thus, summing up all contributions from $t$ to $0$ via backpropagation, we can yield the gradient at the time step $t+1$
\begin{align}
\frac{ \partial \mathcal{L}(t+1) }{ \partial W_{xh} } =\sum_{k=1}^{t+1} \frac{ \partial \mathcal{L}(t+1) }{ \partial {\bf h}_{t+1} }     \frac{ \partial {\bf h}_{t+1}  }{ \partial {\bf h}_{k}  }      \frac{ \partial {\bf h}_{k}  }{  \partial W_{xh}  } 
\end{align}

Further, we can take derivative w.r.t. ${W}_{xh}$ over the whole sequence as
%\begin{align}\label{ch6-eq:rnngradht}
%\frac{ \partial \mathcal{L}(t+1) }{ \partial {\bf h}_{t} } =\frac{ \partial \mathcal{L}(t+1) }{ \partial {\bf h}_{t+1} } \frac{\partial {\bf h}_{t+1} }{ \partial {\bf h}_{t} } +  \frac{ \partial \mathcal{L} }{ \partial z_t } \frac{\partial z_t}{ \partial {\bf h}_{t} }
%\end{align}
\begin{align}\label{ch6-eq:rnngradwxh}
\frac{ \partial \mathcal{L} }{ \partial W_{xh} } = \sum_t \sum_{k=1}^{t+1}  \frac{ \partial \mathcal{L}(t+1) }{  \partial z_{t+1}  }    \frac{  \partial z_{t+1} }{ \partial {\bf h}_{t+1}  }   \frac{  \partial {\bf h}_{t+1} }{ \partial {\bf h}_k  }  \frac{   \partial {\bf h}_{k}  }{ \partial W_{xh} } 
\end{align}

However, there are gradient vanishing or exploding problems to RNNs. Notice that $\frac{  \partial {\bf h}_{t+1} }{ \partial {\bf h}_k  }$ in Eq. \ref{ch6-eq:rnngradwxh} indicates matrix multiplication over the sequence. Because RNNs need to backpropagate gradients over a long sequence (with small values in the matrix multiplication), gradient value will shrink layer over layer, and eventually vanish after a few time steps. Thus, the states that are far away from the current time step does not contribute to the parameters' gradient computing (or parameters that RNNs is learning). Another direction is the gradient exploding, which attributed to large values in matrix multiplication. 
%They have a zero gradient and drive other gradients in previous layers towards 0. Thus, with small values in the matrix and multiple matrix multiplications ( in particular) the gradient values are shrinking exponentially fast, eventually vanishing completely after a few time steps. Gradient contributions from Òfar awayÓ steps become zero, and the state at those steps doesnÕt contribute to what you are learning

Considering the weakness of RNNs, long short term memory (LSTM) was proposed to handle gradient vanishing problem \cite{Hochreiter97}. Notice that RNNs makes use of a simple $tanh$ function to incorporate the correlation between ${\bf x}_t$ and ${\bf h}_{t-1}$ and ${\bf h}_t$, while LSTM model such correlation with a memory unit. And LSTM has attracted great attention for time series data recently and yielded significant improvement over RNNs. For example, LSTM has demonstrated very promising results on handwriting recognition task. In the following part, we will introduce LSTM, which introduces memory cells for the hidden states.%The advantages of deep learning are that they give mappings which can capture meaningful structure information in the code space and introduce bias towards configurations of the parameter space that are helpful for unsupervised learning \cite{Erhan10}. More specifically, it learns the composition of multiple non-linear transformations (such as stacked restricted Boltzmann machines), with the purpose to yield more abstract and ultimately more useful representations \cite{Bengio12}. 
%Moreover, the fine-tuning process can greatly improve performance after the greedy layer-wise unsupervised pre-training \cite{Salakhutdinov09}. %Moreover, a two-step process can greatly improve performance: unsupervised pre-training with greedy layer-wise learning, is then followed by a fine-tuning step in the algorithm \cite{Hinton06b,Salakhutdinov09,Tang13}. %
%In addition, deep learning with gradient descent scales linearly in time and space with the number of train cases, which makes it possible to apply to large scale data sets \cite{Hinton06b}. 

\subsection{Long short-term memory (LSTM)}
\begin{figure*}
\centering
\begin{tabular}{c}
\includegraphics[trim=30 30 30 30, clip, width=10cm]{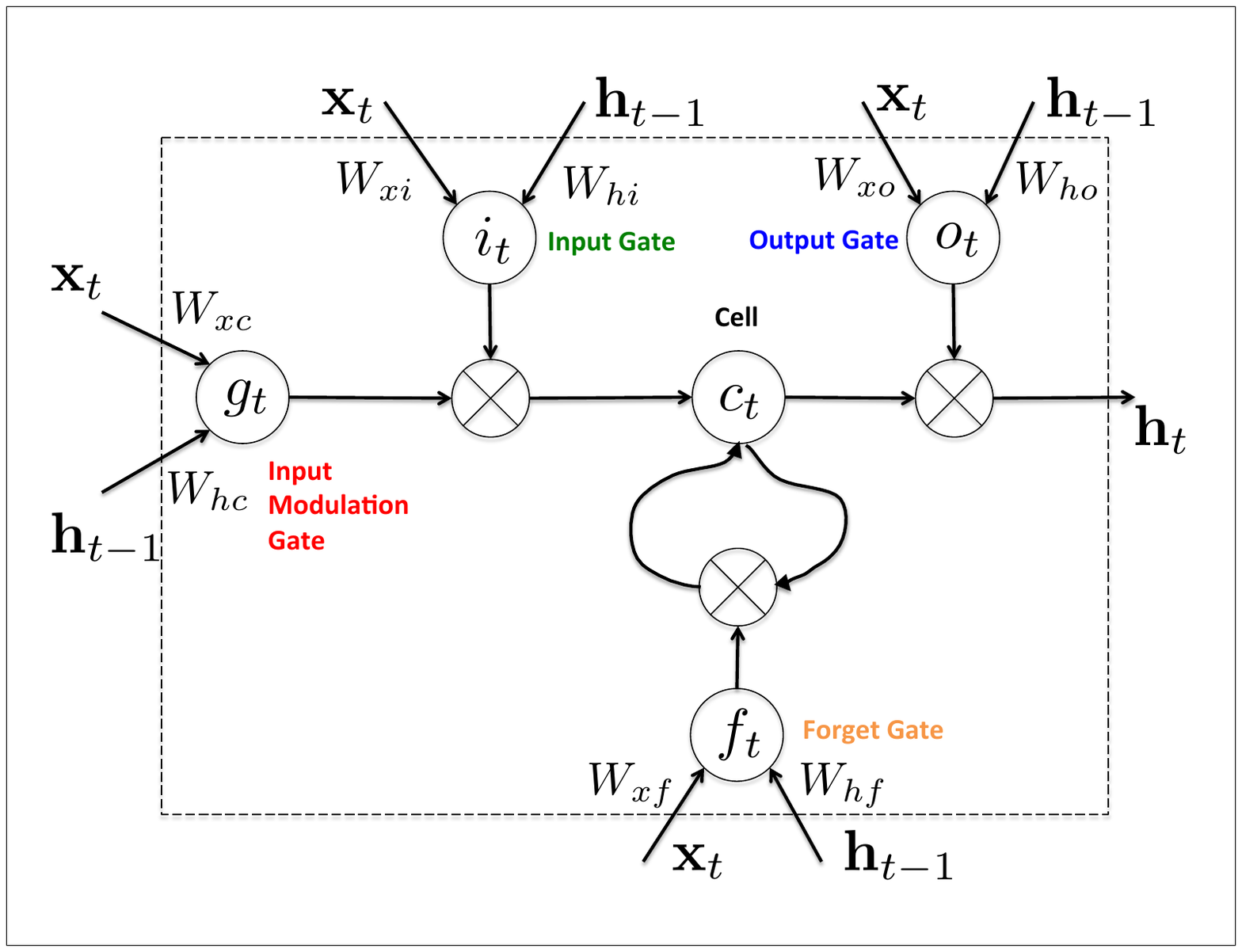} 
\end{tabular}
\caption{It is an unit structure of LSTM, including 4 gates: input modulation gate, input gate, forget gate and output gate.}
\label{ch6-fig:lstm_example}
\end{figure*}

The architecture of RNNs have cycles incorporating the activations from previous trim steps as input to the network to make a decision for the current input, which makes RNNs better suited for sequential labeling tasks. However, one vital problem of RNNs is the gradient vanishing problem. 
LSTM \cite{Hochreiter97} extends the RNNs model with two advantages: (1) introduce the memory information (or cell) (2) handle long sequential better, considering the gradient vanishing problem, refer to Fig. \ref{ch6-fig:lstm_example} for the unit structure. In this part, we will introduce how to forward and backward LSTM neural network, and we will derive gradients and backpropagate error in details.

The core of LSTM is a memory unit (or cell) ${\bf c}_t$ in Fig. \ref{ch6-fig:lstm_example}, which encodes the information of the inputs that have been observed up to that step. The memory cell ${\bf c}_t$ has the same inputs (${\bf h}_{t-1} $ and ${\bf x}_t$) and outputs ${\bf h}_t$ as a normal recurrent network, but has more gating units which control the information flow. The input gate and output gate respectively control the information input to the memory unit and the information output from the unit. More specifically, the output ${\bf h}_t$ of the LSTM cell can be shut off via the output gate.

As to the memory cell itself, it is also controlled with a forget gate, which can reset the memory unit with a sigmoid function. More specifically, given a sequence data $\{ {\bf x}_1,..., {\bf x}_T \}$ we have the gate definition as follows:
\begin{align}
& {\bf f}_t  = \sigma( W_{xf} {\bf x}_t  + W_{hf} {\bf h}_{t-1} + b_f   ) \label{ch6-eq:lstmforgetgate} \\
& {\bf i}_t = \sigma ( W_{xi}{\bf x}_t  + W_{hi} {\bf h}_{t-1}  + b_i    ) \label{ch6-eq:lstminputgate} \\
& {\bf g}_t =  tanh(W_{xc} {\bf x}_t + W_{hc} {\bf h}_{t-1} + b_{c}) \label{ch6-eq:lstmmodulationdgate} \\
& {\bf c}_t = {\bf f}_t \circ {\bf c}_{t-1} + {\bf i}_t \circ {\bf g}_t  \label{ch6-eq:lstmcell} \\
& {\bf o}_t = \sigma(  W_{xo} {\bf x}_t  + W_{ho} {\bf h}_{t-1}  + b_{o}  ) \label{ch6-eq:lstmoutputgate} \\
& {\bf h}_t = {\bf o}_t \circ tanh({\bf c}_t),\ z_t = softmax(W_{hz} {\bf h}_t + b_z) \label{ch6-eq:lstmoutput}
\end{align}
where ${\bf f}_t$ indicates forget gate, ${\bf i}_t$ input gate, ${\bf o}_t$ output gate and ${\bf g}_t$ input modulation gate. Note that the memory unit models much more information than RNNs, except Eq. \ref{ch6-eq:lstmoutputgate}. 

Same as RNNs to make prediction, we can add a linear model over the hidden state ${\bf h}_t$, and output the likelihood with softmax function
\begin{align}
z_t =  softmax(W_{hz} {\bf h}_t + b_z)  \nonumber
\end{align}
If the groundtruth at time $t$ is $y_t$, we can consider minimizing least square $\frac{1}{2}(y_t - z_t)^2$ or cross entroy to estimate model parameters. Thus, for the top layer classification with weight $W_{hz}$, we can take derivative w.r.t. $z_t$ and $W_{hz}$ respectively
\begin{align}
& d{z_t} = y_t - z_t   \label{ch6-eq:lstmtop1}  \\
& d{W_{hz}}  =\sum_t {\bf h}_t    d{ z_t }  \label{ch6-eq:lstmtop2} \\
& d{  {\bf h}_T } =   W_{hz} d{z_T}  \label{ch6-eq:lstmtop3} 
\end{align}
where we only consider the gradient w.r.t. ${\bf h}_T $ at the last time step $T$. For any time step $t$, its gradient will be a little different, which will be introduced later (refer to Eq. \ref{ch6-eq:lstmhiddens2}).

\begin{figure*}
\centering
\begin{tabular}{c}
\includegraphics[trim=50 100 20 100, clip, width=10cm]{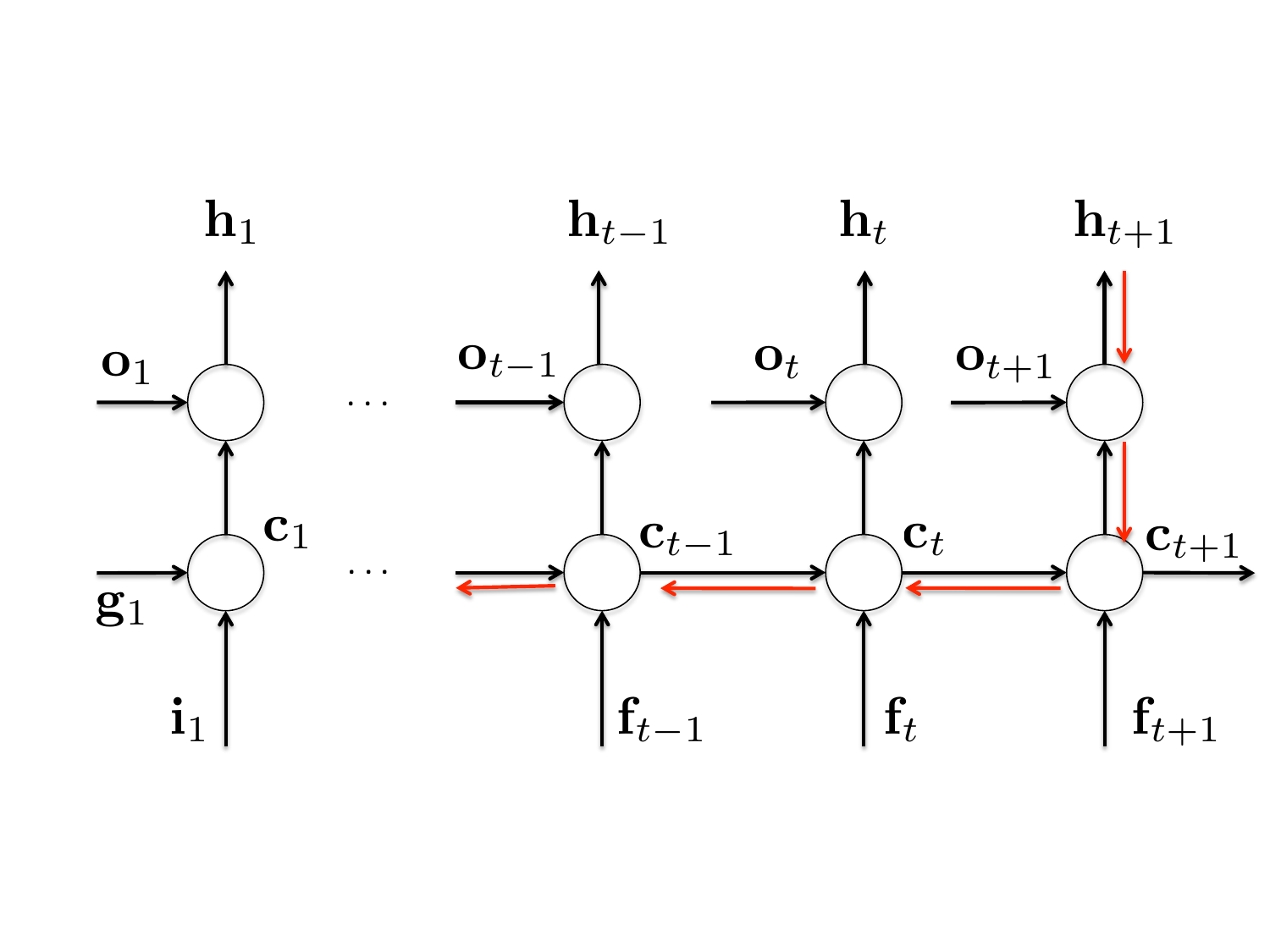} 
\end{tabular}
\caption{This graph unfolds the memory unit of LSTM, with the purpose to make it easy to understand error backpropagation.}
\label{ch6-fig:lstm_unfold}
\end{figure*}

then backpropagate the LSTM at the current time step $t$
\begin{align}
& d{{\bf o}_t}  = tanh({\bf c}_t) d {\bf h}_t  \label{ch6-eq:lstmcell1} \\
& d {{\bf c}_t} =  \big(1- tanh( {\bf c}_t)^2 \big)  {\bf o}_t  d {\bf h}_t   \label{ch6-eq:lstmcell2} \\
& d {{\bf f}_t} ={\bf c}_{t-1}   d {{\bf c}_t}  \label{ch6-eq:lstmcell3} \\
& d{ {\bf c}_{t -1}} \  +=\  {\bf f}_t  \circ d {\bf c}_{t} \label{ch6-eq:lstmcell4} \\
& d { {\bf i}_t}  = {\bf g}_t   d{{\bf c}_t} \label{ch6-eq:lstmcell5} \\
& d {{\bf g}_t}  = {\bf i}_t d{ {\bf c}_t}  \label{ch6-eq:lstmcell6}
\end{align}
where the gradient w.r.t. $tanh({\bf c}_t)$ in Eq. \ref{ch6-eq:lstmcell2} can be derived according to Eq. \ref{ch6-eq:tanhgradient} in Appendix.

further, back-propagate activation functions over the whole sequence
\begin{align}
& d{ W_{xo} } = \sum_t  {\bf o}_t (1 - {\bf o}_t) {\bf x}_t d{{\bf o}_t}  \nonumber \\
& d{ W_{xi} }  = \sum_t  {\bf i}_t (1 -{\bf i}_t) {\bf x}_t  d{{\bf i}_t}  \nonumber \\
& d{  W_{xf} }  = \sum_t {\bf f}_t(1 - {\bf f}_t) {\bf x}_t d{{\bf f}_t} \nonumber \\
& d{ W_{xc} } = \sum_t ( 1-{\bf  g}_t^2 ) {\bf x}_t d{{\bf g}_t} \label{ch6-eq:lstmactivation}
\end{align}
Note that the weights $W_{xo}$, $W_{xi}$, $W_{xf}$ and $W_{xc}$ are shared across the whole sequence, thus we need to take the same summation over $t$ as RNNs in Eq. \ref{ch6-eq:rnngradwhz}.
Similarly, we have
\begin{align}
& d{ W_{ho} } =\sum_t  {\bf o}_t(1 - {\bf o}_t)  {\bf h}_{t-1} d{{\bf o}_t}  \nonumber \\
& d{ W_{hi}  }  =\sum_t  {\bf i}_t (1 - {\bf i}_t) {\bf h}_{t-1}  d{{\bf i}_t}  \nonumber \\
& d{  W_{hf} }  = \sum_t {\bf f}_t(1 - {\bf f}_t) {\bf h}_{t-1} d{{\bf f}_t} \nonumber \\
& d{ W_{hc} } = \sum_t ( 1- {\bf g}_t^2 ) {\bf h}_{t-1} d{{\bf g}_t} \label{ch6-eq:lstmhiddenweights}
\end{align}

and corresponding hiddens at the current time step $t-1$\\
\begin{align}
d{{\bf h}_{t-1}} & = {\bf o}_t (1 - {\bf o}_t)  W_{ho} d{{\bf o}_t}  +  {\bf i}_t (1 -{\bf  i}_t) W_{hi}  d{{\bf i}_t}  \nonumber \\
&   +   {\bf f}_t(1 - {\bf f}_t)  W_{hf} d{{\bf f}_t}  +  ( 1- {\bf g}_t^2 )  W_{hc}  d{{\bf g}_t}   \label{ch6-eq:lstmhiddens1} \\
d{ {\bf h}_{t - 1} } & = d{{\bf h}_{t-1}}  +      W_{hz} d{z_{t-1}}  \label{ch6-eq:lstmhiddens2}
\end{align}
where we consider two sources to derive $d{{\bf h}_{t-1}}$, one is from activation function in Eq. \ref{ch6-eq:lstmhiddens1} and the other is from the objective function at the time step $t-1$ in Eq. \ref{ch6-eq:lstmtop3}.
%\subsection{Bi-directional long short term memory (LSTM)}

\subsection{Error backpropagation}
In this part, we will give detail information on how to derive gradients, especially Eq. \ref{ch6-eq:lstmcell4}. As we talked about RNNs before, we can take the same strategy to unfold the memory unit, shown in Fig. \ref{ch6-fig:lstm_unfold}. Suppose we have the least square objective function 
\begin{align}
\mathcal{L}({\bf x}, \theta) = \min \sum_t \frac{1}{2} (y_t - z_t)^2 
\end{align}
where $\boldsymbol{\theta} = \{ W_{hz},  W_{xo}, W_{xi}, W_{xf}, W_{xc} , W_{ho}, W_{hi}, W_{hf}, W_{hc} \}$ with biases ignored. To make it easy to understand in the following, we use $\mathcal{L}(t) = \frac{1}{2} (y_t - z_t)^2$. 

At the time step $T$, we take derivative w.r.t. ${\bf c}_{T}$
\begin{align}
\frac{\partial \mathcal{L}(T)  }{ \partial {\bf c}_{T}  } = \frac{  \partial \mathcal{L}(T)   }{  \partial {\bf h}_{T}  }    \frac{ \partial  {\bf h}_{T}  }{\partial {\bf c}_{T}  }
\end{align}

At the time step $T-1$, we take derivative of $\mathcal{L}(T-1)$ w.r.t. ${\bf c}_{T-1}$ as 
\begin{align}
\frac{\partial \mathcal{L}(T-1)  }{ \partial {\bf c}_{T-1}  } = \frac{  \partial \mathcal{L}(T-1)   }{  \partial {\bf h}_{T-1}  }    \frac{ \partial  {\bf h}_{T-1}  }{\partial {\bf c}_{T-1}  }
\end{align}
However, according to Fig. \ref{ch6-fig:lstm_unfold}, the error is not only backpropagated via $\mathcal{L}(T-1)$, but also from ${\bf c}_{T}$, thus the final gradient w.r.t. ${\bf c}_{T-1}$ 
\begin{align}
& \frac{\partial \mathcal{L}(T-1)  }{ \partial {\bf c}_{T-1}  } = \frac{  \partial \mathcal{L}(T-1)   }{  \partial  {\bf c}_{T-1} }    + \frac{\partial \mathcal{L}(T)  }{ \partial {\bf c}_{T-1}  }  \nonumber \\
& \frac{\partial \mathcal{L}(T-1)  }{ \partial {\bf c}_{T-1}  } = \frac{  \partial \mathcal{L}(T-1)   }{  \partial {\bf h}_{T-1}  }    \frac{ \partial  {\bf h}_{T-1}  }{\partial {\bf c}_{T-1}  } + \frac{  \partial \mathcal{L}(T)   }{  \partial {\bf h}_{T}  }    \frac{ \partial  {\bf h}_{T}  }{ \partial {\bf c}_{T}  }    \frac{ \partial {\bf c}_{T}  }{  \partial {\bf c}_{T-1}  }  \label{ch6-eq:lstmgradientchain}
\end{align}
where we use the chain rule in Eq. \ref{ch6-eq:lstmgradientchain}. Further, we can rewrite Eq. \ref{ch6-eq:lstmgradientchain} as
\begin{align}
d{ {\bf c}_{T-1}} \  =  d{ {\bf c}_{T-1}}  +  {\bf f}_T \circ d {\bf c}_{T}  
\end{align}
In a similar manner, we can derive Eq. \ref{ch6-eq:lstmcell4} at any time step.

\subsection{Parameters learning}
{\bf Forward}: we can use Eqs. \ref{ch6-eq:lstmforgetgate}-\ref{ch6-eq:lstmoutput} to update states as the feedforward neural network from the time step $1$ to $T$.

{\bf Backword}: we can backpropagate the error from $T$ to $1$ via Eqs. \ref{ch6-eq:lstmcell1}-\ref{ch6-eq:lstmhiddens2}. After we get gradients using backpropagation, the model $\boldsymbol{\theta}$ can be learnt with gradient based methods, such as stochastic gradient descent and L-BFGS). If we use stochastic gradient descent (SGD) to update $\boldsymbol{\theta} = \{ W_{hz},  W_{xo}, W_{xi}, W_{xf}, W_{xc} , W_{ho}, W_{hi}, W_{hf}, W_{hc} \}$, then we have
\begin{equation}
\boldsymbol{\theta} = \boldsymbol{\theta} - \eta d{ \boldsymbol{\theta}  }
%\Theta = \Theta - \eta \frac{\partial \mathrm{log} \mathcal{L}_{dis}(\mathcal{D};\Theta)}{\partial \Theta})
\label{ch6-eq:grbm}
\end{equation}
where $\eta$ is the learning rate. Notice that more tricks can be applied here.
\section{Applications}
RNNs can be used to handle sequential data, such as speech recognition. RNNs can also be extended into multiple layers in a bi-directional manner. Moreover, RNNs can be combines with other neural networks \cite{Chen15}, such as convolutional neural network to handle video to text problem, as well as combining two LSTMs for machine translation. 

\section{Appendix}

(1) Gradients related to $f(x) = tanh(x)$\\
Take the derivative of $tanh(x)$ w.r.t. $x$
\begin{align}\label{ch6-eq:tanhgradient}
& \frac{ \partial{tanh(x)} }{ \partial(x)  }  =  \frac{ \partial{\frac{ sinh(x) }{ cosh(x) } }  }{  \partial{x} }  \nonumber \\
=  &\frac{  \frac{ \partial{ sinh(x)}  }{ \partial{x} }  cosh(x) - sinh(x) \frac{ \partial{ cosh(x) }}{ \partial{x} }     }{ \big( cosh(x) \big)^2 }  \nonumber \\
= & \frac{[ cosh(x) ]^2 -[ sinh(x)]^2 }{  \big( cosh(x) \big)^2  } \nonumber \\
= & 1 - [tanh(x)]^2
\end{align}

(2) Gradients related to Eq. \ref{ch6-eq:rnngradz}\\
As we know $z = softmax(W_{hz} {\bf h} + {\bf b}_z)$ predicts the probability assigned to $K$ classes without considering the time step information in Eq. \ref{ch6-eq:rnngradz}. Furthermore, we can use $1$ of $K$ encoding to represent the groundtruth $y$, but with probability vector to represent $z =  [p(\hat{y}_1),...,p(\hat{y}_K) ]$. Then, we can consider the gradient in each dimension, and then generalize it to the vector case in the objective function $\mathcal{L}(W_{hz}, {\bf b}_z) = - y log z$. In the following, we will first compute the gradient w.r.t. $\alpha_j(\Theta) =  W_{hz}(:, j)  {\bf h}_t $, and then generalize it to $k\neq j$. And further, we can derive the gradients w.r.t. $W_{hz}$ and ${\bf b}_z$. 

We know that 
\begin{align}\label{ch6-eq:softmax}
p(\hat{y}_j | {\bf h}_t; \Theta)   = \frac { \textrm{exp} ( \alpha_j (\Theta) ) }{\sum_{k}  \textrm{exp} (  \alpha_k (\Theta)   ) }
\end{align}
Then take the derivative w.r.t. $\alpha_j(\Theta)$
\begin{align}\label{ch6-eq:softmaxgrad1}
&\frac{ \partial{y_j \textrm{log} p(\hat{y}_j | {\bf h}_t; \Theta) } }{   \partial \alpha_j}  \nonumber \\
& =\frac{y_j}{ p(\hat{y}_j) }  \frac { \textrm{exp} ( \alpha_j (\Theta) ) \sum_{k}  \textrm{exp} (  \alpha_k (\Theta)   )  -  \textrm{exp} ( \alpha_j (\Theta) ) \textrm{exp} ( \alpha_j (\Theta) ) }{ [\sum_{k}  \textrm{exp} (  \alpha_k (\Theta)   ) ]^2 }  \nonumber  \\
& = y_j (1 - p(\hat{y}_j))
\end{align}

Similarly, $\forall k\neq j$ and its prediction $p(\hat{y}_k)$, we take the derivative w.r.t. $\alpha_j(\Theta)$, 
\begin{align}\label{ch6-eq:softmaxgrad2}
& \frac { \partial{y_k \textrm{log} p(\hat{y}_k | {\bf h}_t; \Theta)  }}{   \partial \alpha_j}  \nonumber  \\
& = \frac{y_k}{p(\hat{y}_k)} \frac { -  \textrm{exp} ( \alpha_k (\Theta) ) \textrm{exp} ( \alpha_j (\Theta) ) }{[\sum_{s}  \textrm{exp} (  \alpha_s (\Theta)   )]^2 }  \nonumber  \\
& = - y_k p(\hat{y}_j) 
\end{align}
Finally, we can yield the following gradient w.r.t. $\alpha_j(\Theta)$ 
\begin{align}\label{ch6-eq:softmaxgrad}
\frac{\partial p(\hat{\bf y}) }{  \partial \alpha_j }  & =  \sum_j \frac{\partial{ y_j \textrm{log} p(\hat{y}_j | {\bf h}_i; \Theta) } }{   \partial \alpha_j}   \nonumber \\
& = \frac{\partial{ \textrm{log} p(\hat{y}_j | {\bf h}_i; \Theta) } }{   \partial \alpha_j} + \sum_{k \neq j} \frac {\partial{ \textrm{log} 
p(\hat{y}_k | {\bf h}_i; \Theta)  }}{   \partial \alpha_j}    \nonumber  \\
& = y_j - y_j p(\hat{y}_j)  - \sum_{k \neq j }  y_k  p(\hat{y}_j)    \nonumber  \\
& = y_j - p(\hat{y}_j) (y_j + \sum_{k \neq j } y_k ) = y_j - p(\hat{y}_j) 
\end{align}
where we use the results in Eqs. \ref{ch6-eq:softmaxgrad1} and \ref{ch6-eq:softmaxgrad2}.

\end{document}